\documentclass{article}

\usepackage[dblblindworkshop, final]{neurips_2025}

\usepackage[utf8]{inputenc} 
\usepackage[T1]{fontenc}    
\usepackage{hyperref}       
\usepackage{url}            
\usepackage{booktabs}       
\usepackage{amsfonts}       
\usepackage{nicefrac}       
\usepackage{microtype}      
\usepackage{xcolor}         
\definecolor{darkblue}{rgb}{0,0,.4}
\usepackage{natbib}         
\usepackage{amssymb}
\usepackage{graphicx}
\graphicspath{ {./images/} }

\hypersetup{
    colorlinks=true,
    linkcolor=darkblue,
    citecolor=darkblue,
    filecolor=magenta,      
    urlcolor=cyan
}

\title{Open-weight genome language model safeguards: Assessing robustness via adversarial fine-tuning}
\workshoptitle{Biosecurity Safeguards for Generative AI}

\author{%
James R.\ M.\ Black\\
Center for Health Security\\
Johns Hopkins Bloomberg School of Public Health
\And
Moritz S.\ Hanke\\
Center for Health Security\\
Johns Hopkins Bloomberg School of Public Health
\And
Aaron Maiwald\\
Department of Chemistry\\
University of Oxford
\And
Tina Hernandez\mbox{-}Boussard\\
Stanford University School of Medicine
\And
Oliver M. Crook\\
Department of Chemistry \& Kavli Institute for Nanoscience Discovery\\
University of Oxford
\And
Jaspreet Pannu\footnotemark[1]\\
Center for Health Security\\
Johns Hopkins Bloomberg School of Public Health\\
\texttt{pannu@jhu.edu}}

\begin{document}
\maketitle
\begin{abstract}
Genomic language models (gLMs) have demonstrated impressive predictive and generative capabilities, raising concerns that such models may also enable misuse, for instance via the generation of genomes for human-infecting viruses. These concerns have catalyzed calls for risk mitigation measures. The de facto mitigation of choice is filtering of pretraining data (i.e., removing viral genomic sequences from training datasets) in order to limit gLM performance on virus-related tasks. However, it is not currently known how robust this approach is for securing open-source models that can be fine-tuned using sensitive pathogen data. Here, we evaluate a state-of-the-art gLM, Evo 2, and perform fine-tuning using sequences from 110 harmful human-infecting viruses to assess the rescue of misuse-relevant predictive capabilities. The fine-tuned model exhibited reduced perplexity on unseen viral sequences relative to the pretrained model. The model fine-tuned on human-infecting viruses also showed signs of identifying immune escape variants from SARS-CoV-2 (achieving an AUROC of 0.59), despite having no exposure to SARS-CoV-2 sequences during fine-tuning. This work demonstrates that data exclusion can be partially circumvented by fine-tuning approaches that can, to some degree, rescue misuse-relevant capabilities of gLMs. We highlight the need for safety frameworks for gLMs and outline further work needed on evaluations and mitigation measures to enable the safe deployment of gLMs. 
\end{abstract}

\section{Introduction}
\setlength{\parskip}{0pt} 
\setlength{\parindent}{1em} 
Large language models (LLMs) have transformed our ability to collect and generate information encoded in human language, and today match or outperform humans on a multitude of complex tasks \citep{zhao2025surveylargelanguagemodels}. Deep learning methodologies originally developed for LLMs, such as self-supervised learning on large amounts of unlabeled data, have been applied to biological data such as protein and genetic sequences. These models are referred to as protein language models (pLMs) or genomic language models (gLMs), respectively \citep{benegas_genomic_2024}. Early pLMs and gLMs have demonstrated intriguing capabilities, such as the prediction of protein structural elements and transcription factor binding sites \citep{chowdhury_single-sequence_2022, mendoza-revilla_foundational_2024}. The performance limits of these models is not yet clear and may be gated by the availability of high-quality training data, compute, or novel AI architectures particularly well-suited to biological data. Thus, the potential ceiling performance of biological AI models may be contingent on future, adjacent advances in computational biotechnology.

Scientists have noted that to the degree these systems acquire novel biological capabilities, they might also be vulnerable to misuse \citep{committee_on_assessing_and_navigating_biosecurity_concerns_and_benefits_of_artificial_intelligence_use_in_the_life_sciences_age_2025}. Developers have acknowledged these risks and taken first steps towards mitigating gLM and pLM risks. However, given the uncertainties regarding the upper limit of model performance described above, it is challenging to determine the best approaches for mitigating misuse risks, both for today and for years into the future. Developers have acknowledged these risks and taken first steps towards mitigating gLM and pLM risks. A handful of risk mitigation approaches have been tested to date. For instance, the developers of ESM3-open "removed the capability for the model to follow prompts related to viruses and toxins" for their open-source model ESM3-open \citep{hayes_simulating_2025}. 

The de facto mitigation of choice for these models, though, is known as data exclusion (also known as data filtering). Data exclusion involves the deliberate removal of data from training datasets in order to limit model performance on risky capabilities tied to such data, such as capabilities related to biological weapons development. For safety and security reasons, the developers of the Evo models, a series of gLMs “excluded genomic sequences from viruses that infect eukaryotic hosts (...) to ensure our openly shared model did not disseminate the capability to manipulate and design pathogenic human viruses” \citep{brixi_genome_2025}. Similarly, for ESM3-open, the developers "removed sequences unique to viruses, as well as viral and non-viral sequences from the Select Agents and Toxins List" from the training data in order "to reduce the capability of ESM3-open on these sequences" \citep{hayes_simulating_2025}.

However, relatively little is known about the true performance reduction that results from the exclusion of sensitive data. Some researchers have demonstrated that this risk reduction may not be robust \citep{zhang_genebreaker_2025}. Open-source models can be fine-tuned after their release, suggesting that a sufficiently skilled actor could rescue model performance for misuse-related tasks if the sensitive data in question were available to them. The quantity of sensitive data required in this context to rescue the misuse-enabling capability, as well as the degree to which those capabilities can be rescued, has not yet been tested. Further, some researchers have suggested that highly capable biological models could interpolate risky capabilities due to their broad generalizability, even when particular subsets of training data have been removed. Such generalizability for gLMs might exist analogously to how it was demonstrated for protein structure prediction capabilities \citep{ahdritz_openfold_2024}.

Here, we conduct evaluations on the robustness of data exclusion from the open-source gLM, Evo 2, by reintroducing sequences from human-infecting viruses via fine-tuning and assessing the rescue of misuse-relevant capabilities. Generally, Evo 2 learns patterns from DNA sequences to both 1) predict the impact of genetic sequence changes and 2) generate genetic sequences; we evaluate primarily the predictive capability in its application to viruses. We demonstrate that important misuse-relevant capabilities for viruses unseen in the training data, including the prediction of downstream phenotypes related to immune escape, can be rescued through adversarial fine-tuning. Lastly, we discuss frameworks for systematically approaching safety and security considerations for gLMs to improve biosecurity in the future.

\section{Methods}

Note: Due to concerns about information hazards arising from this work, some details have been redacted from the Methods section, and code pertaining to this study has not been made available but can be obtained from the authors by reasonable request.

\subsection{Base model used for fine-tuning}

We used the gLM Evo 2 \citep{brixi_genome_2025}, an open-source model available in three versions with 1 billion (1B), 7 billion (7B), and 40 billion (40B) parameters, respectively. Evo 2-7B  was used as the base model for fine-tuning. 

\subsection{Fine-tuning methodology}

Fine-tuning was performed using the Evo 2-7B  as a base, leveraging the Savanna framework for distributed training. Briefly, this approach uses a transformer backbone with rotary position embeddings and was configured for autoregressive pretraining. The model was distributed across 4 NVIDIA H100 GPUs using DeepSpeed ZeRO Stage 1 optimization with mixed precision (BF16) training. The pretrained Evo 2 7B weights were fine-tuned using a character-level tokenizer, with the model adapted from its original 1M context length to 4,096 tokens per sample due to computational constraints. Training utilized the Adam optimizer ($\beta 1=0.9$, $\beta 2=0.95$, $lr=4x10e-6$), with an effective batch size of 144 sequences. We applied gradient clipping (0.5), weight decay ($1x10e-4$), and dropout (0.05) for stability. Training was conducted for 150 steps with a cosine learning rate schedule, decaying from the peak learning rate to a minimum of 8x10e-8, with 10\% warmup.

\subsection{Sequence data used for fine-tuning}

The dataset used for adversarial fine-tuning consisted of viral sequences. We attempted to collate this dataset in an analogous way to the approach taken by the Evo 2 authors in collating their OpenGenome2 dataset, which was utilized for training in the first instance. First, a literature search was performed for putative harmful human-infecting viruses across six groups of concern: large DNA viruses, small DNA viruses, positive-strand RNA viruses, negative-strand RNA viruses, enteric viruses, and double-stranded viruses. Second, deduplication was performed in a manner that mimicked the approach taken by the Evo 2 team, in order to remove redundancy while preserving sequence diversity. Complete sequences were deduplicated using Mash sketching with 10,000 k-mers to calculate pairwise distances between all genomes. Sequences with Mash distance <0.01 (i.e., >99 percent average nucleotide identity) were clustered, and the longest sequence from each cluster was considered to be representative \citep{ondov_mash_2016}. This resulted in a dataset of 122 viral genomes (Figure 1). This was then separated using a 90/10 split into a training dataset of 110 viral genomes and a held-out test dataset of 12 viral genomes for downstream evaluation. In parallel, a second, control dataset of prokaryote-infecting viruses (bacteriophages) was generated for a separate fine-tuning exercise, using an analogous approach. This resulted in the generation of 181 bacteriophage genomes.

\begin{figure}[htbp]
\centering
\includegraphics[scale=0.125]{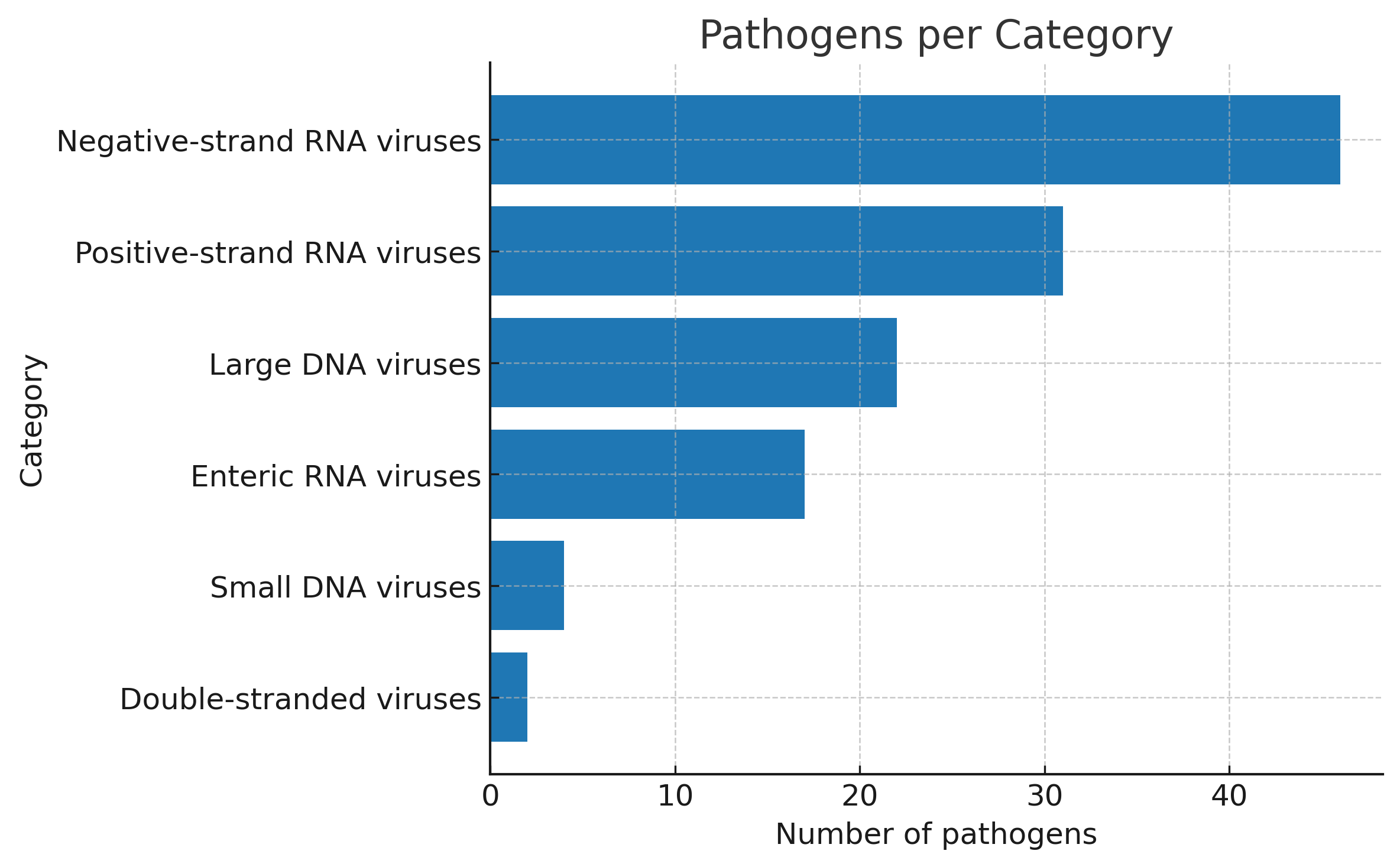}
\caption{Composition of the dataset of viral genomes, by type of viral sequence. 110 viral sequences were used for fine-tuning, and 12 sequences were held out for downstream evaluation.}
\label{fig:dataset-composition}
\end{figure}

\subsection{Evaluations}

We evaluated two primary capabilities, relating to the predictive, rather than the generative, ability of Evo 2. 

Firstly, sequence prediction (perplexity): We evaluated sequence perplexity on three different versions of Evo 2: the pretrained model; the control version, fine-tuned on bacteriophage sequences (FT-bacteriophages); and the version fine-tuned on harmful human-infecting viruses (FT-harmful). Perplexity was calculated separately on the training data, and the held-out set of 12 viral genomes.  

Secondly, phenotype prediction: We assessed the ability of the three different versions of Evo 2 to predict which mutations to the SARS-CoV-2 spike protein would underpin immune escape. Importantly, none of the three models had seen any SARS-CoV-2 genomes at any stage of training or fine-tuning. Immune escape was determined using publicly available deep mutational scanning data, leveraging polyclonal or monoclonal antibodies from infected or vaccinated individuals \citep{starr_deep_2020}. Mutations were divided into those that were likely to underpin immune escape and those that were not likely to underpin immune escape. Sequence perplexity was computed for the sequence encoding the spike protein. The ability of the three models to identify immune escape mutations was compared to BLOSUM-62 scores, representing simple heuristics of evolutionary conservation, and EVEscape, a highly specialized predictor based on a deep learning approach that combines temporal sequences with predicted structural and biophysical information, which was considered the gold standard for this task \citep{thadani_learning_2023}.

\section{Results}

\subsection{Perplexity on held-out viral sequences}
In order to understand the degree to which exclusion of sensitive data was a robust mitigation measure for a gLM, we first assessed whether fine-tuning on that data might rescue model performance related to viral sequences. Perplexity provides an approach for assessing this, measuring how well a gLM predicts the next token in a sequence, with lower values indicating better predictive performance and thus a better understanding of the underlying genomic data. In the Evo2 paper, the authors demonstrated that perplexity was increased on eukaryote-infecting virus sequences relative to prokaryote-infecting viruses, reflecting their removal from the training data. 

We compared three versions of the 7B parameter model, pretrained Evo 2, and two fine-tuned models, one fine-tuned on the bacteriophage genomes (FT-bacteriophages) and the other on genomes from harmful, human-infecting viruses (FT-harmful). Each version reflected a modified version of the original model, featuring a 1M token context window, with the context window curtailed to 4096 tokens. We evaluated model performance by its ability to predict sequences from within its fine-tuning distribution, as well as a held-out set of viral genomes that had not been included in the fine-tuning dataset. The held-out viruses were balanced across viral families in a similar way to the viruses used for fine-tuning to ensure comprehensive coverage across sequences of interest. 

First, we confirmed that sequence length did not relate to perplexity. This was particularly important as we had implemented post-hoc changes to the model to modify context length. There was no relationship between sequence length and perplexity in either the in-domain or out-of-domain (held-out) datasets (Figure 2). 

\begin{figure}[htbp]
\centering
\includegraphics[scale=0.71]{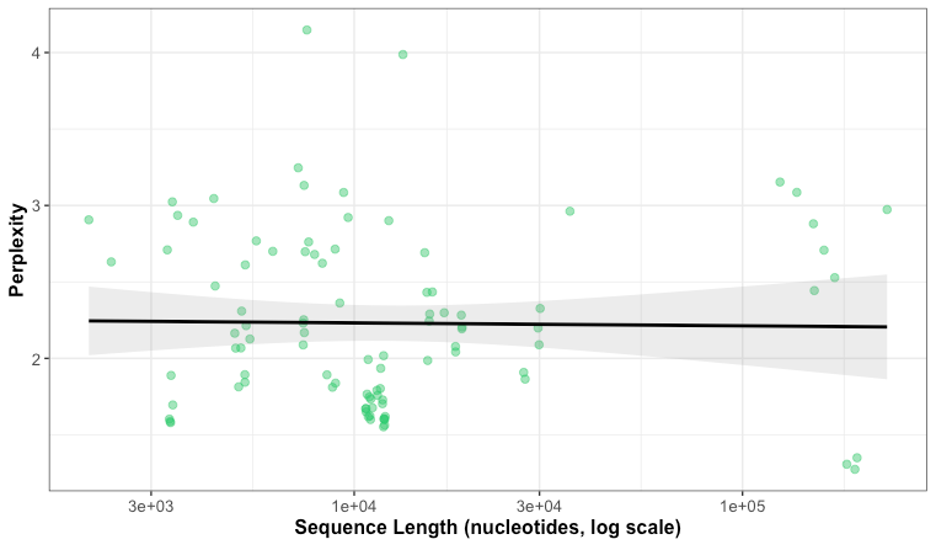}
\caption{Scatterplot showing perplexity and sequence length for 110 harmful human-infecting viruses that the fine-tuned model was trained on (n=97, r=0.034). Perplexity measures how well a gLM predicts the next token in a sequence, with lower values indicating better predictive performance and thus a better understanding of the underlying genomic data. Each point on the plot corresponds to a single virus used for fine tuning.}
\label{fig:scatterplot}
\end{figure}

The FT-harmful model exhibited substantially reduced perplexity, relative to the pretrained model and the FT-bacteriophages model on the training data used for fine-tuning. For the test data, the FT-harmful model exhibited the same trends, although the difference in perplexity observed between FT-bacteriophages and FT-harmful was not statistically significant (Figure 3, training data pretrained: median 3.84; training data FT-bacteriophage: median 3.73; training data FT-harmful: median 2.16; test data pretrained: median 3.83; test data FT-bacteriophage: median 3.73; test data FT-harmful: median 3.55). However, relevant to security concerns, this highlights that the model may have begun to learn generalisable patterns from a limited set of examples. The observation that the pretrained model performed slightly worse than the bacteriophage fine-tuned model might plausibly reflect the damage sustained by the model after the reduction in context length.

\begin{figure}[htbp]
\centering
\includegraphics[scale=0.52]{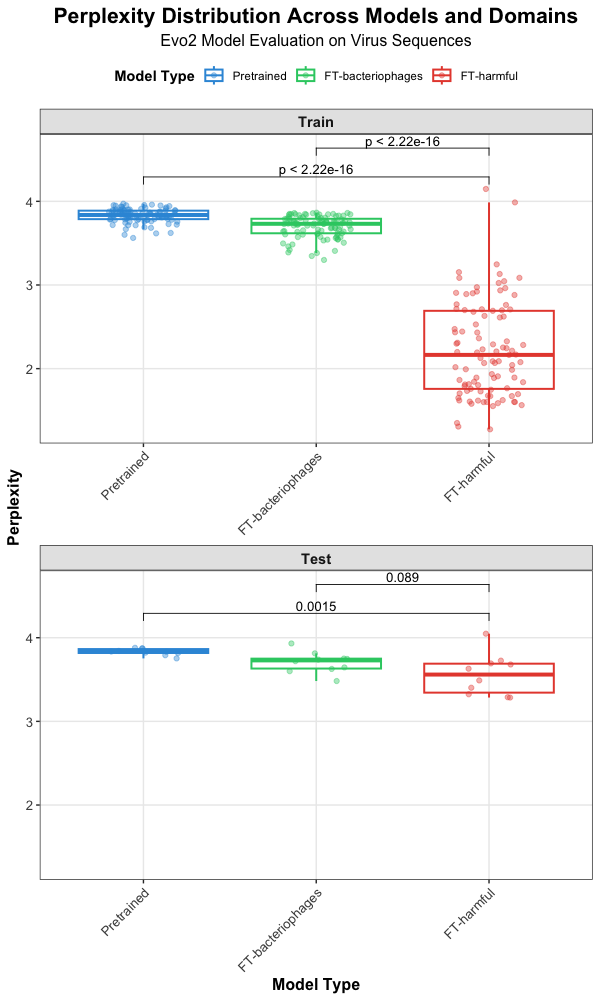}
\caption{Boxplot showing perplexity on training (n=110) and test (n=12) sequences across the three versions of Evo 2: pretrained; FT-bacteriophages; FT-harmful.}
\label{fig:domain}
\end{figure}

\subsection{Prediction of SARS-CoV-2 immune escape}
gLMs might learn patterns within genomic data that correspond to downstream phenotypes. In the context of viral data, these might include transmissibility, virulence, or immune evasion. As an exemplar, we wondered whether the version of Evo 2 fine-tuned on harmful human-infecting viruses might be able to predict which mutations might confer immune evasion for an unseen virus, i.e., not included in the fine-tuning dataset.

To test this, we computed perplexity on the spike protein from the Wuhan SARS-CoV-2 genome, and compared these scores to functional data that has been gathered about this protein relating to host protein interactions and immune escape. Importantly, no SARS-CoV-2 species had been included in the data used for fine-tuning, and so any predictions made by the model could plausibly be considered analogous to performance on an unseen virus. This would represent a relatively stringent level of generalizability, as the model would have to transfer patterns from other viruses to make predictions about a novel pathogen. We leveraged deep mutational scanning data as a baseline, focusing on immune escape due to its clear biosecurity relevance and the immediate availability of high-quality functional data for validation. 

We evaluated the performance of five different approaches at predicting which mutations would lead to immune escape and which would not. We leveraged the AUROC to compare the different approaches. Neither the pretrained model nor the model fine-tuned on bacteriophage data conferred any improved predictive performance relative to a random classifier (Figure 3). In addition, we used the BLOSUM-62 matrix to evaluate the degree to which evolutionary conservation might predict mutations that conferred immune escape. This delivered an AUC of 0.51, in essence, no better than a random classifier. The purpose-built, deep learning tool EVEscape, which can be seen as a gold standard here, conferred an AUC of 0.8.

The version of the model fine-tuned on harmful, human-infecting viruses had an AUC of 0.59 at this task. This suggests that fine-tuning on sensitive data conferred generalized predictive properties related to unseen harmful pathogens, relative to both the baseline model and the version of the model fine-tuned on bacteriophages. However, the performance did not match that of the narrow, purpose-built model EVEScape.

Overall, these results show that training data exclusion is unlikely to be a fully robust method for preventing the emergence of misuse-relevant capabilities from open-source gLMs and potential associated downstream harms.

\begin{figure}[htbp]
\centering
\includegraphics[scale=0.525]{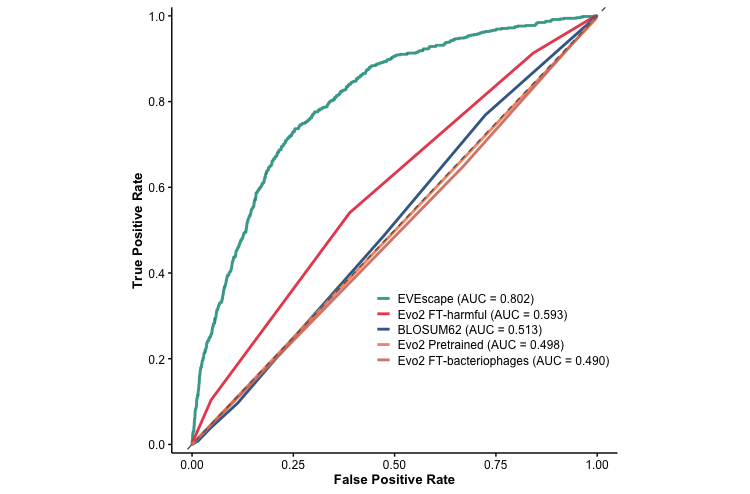}
\caption{Receiver operating characteristic (ROC) curves comparing methods for identifying SARS-CoV-2 Spike mutations leading to a phenotype of immune escape vs no immune escape. 3 versions of Evo 2, pretrained, fine-tuned (bacteriophages), and fine-tuned (harmful human-infecting viruses) were compared. BLOSUM-62 scores were used to evaluate whether evolutionary conservation alone would confer predictive power. EVEscape, a deep learning model leveraging fitness predictions and structural information, was compared as an example of a model specialized for this exact task.}
\label{fig:roc-curves}
\end{figure}

\section{Discussion}

\subsection{Key findings}

We have shown that data exclusion can be circumvented through fine-tuning with publicly available sequence data of human-infecting viruses. We also demonstrated a proof-of-principle that misuse-relevant capabilities of openly available gLMs can be rescued through this process. However, model performance on tasks involving functional immune escape prediction was inferior to a more narrow, purpose-built tool. This is congruent with recent results showing narrow, purpose-built tools can outperform large biological foundation models on many tasks related to viruses \citep{gurev_variant_2025}. Of note, a single fine-tuned gLM might be applicable to a much wider range of misuse-relevant tasks not tested in this study. 

\subsection{Limitations}

Certain limitations imply caution when interpreting our results. Firstly, computational constraints prevented us from working with a full 1M token context model. As it was necessary to modify the model architecture to work with a 4096 token context prior to fine-tuning, it is possible that the capabilities we demonstrate underestimate capabilities that might be achievable if the starting point were the 1M token model instead. Nonetheless, many nefarious actors might face similar constraints. Secondly, whilst we have highlighted example tasks, including predictions of downstream functional data, these do not represent the full spectrum of misuse-relevant gLM capabilities. Ideally, an evaluation of sequence generation properties would have been conducted as well, and future work should address this.

\subsection{Accident and misuse harms linked to gLMs}
A robust evaluation framework should consider the multitude of ways a gLM could be misused. Misuse-enabling capabilities that are most likely or most severe require further investigation first. The most discussed risk to date has been the embedding of functional information regarding pathogens, such as virulence, transmissibility, latency, or immune escape \citep{pannu_dual-use_2025}. Whilst perplexity, downstream functional data, and sequence design evaluations are all relevant, future work should explore other less obvious possibilities, such as gLMs being used to predict the fitness of putative novel or heavily modified viral designs, or the generative properties of gLMs being leveraged in similar steps. Experimental work to validate predicted downstream properties should also be considered, insofar as they are conducted in a secure environment using safe biological proxies \citep{ikonomova_experimental_2025}.

Given that misuse has a much higher tolerance for failure than benign usage, i.e., the release of an ineffective biological agent has fewer negative consequences than the release of an ineffective countermeasure, it is not clear how well-suited evaluations that simply assess perplexity as an objective measure are. For example, it is possible that a model with very high perplexity, but that has a small chance to generate a viable viral design that is significantly different from existing sequences, for instance, with a totally foreign antigen, might be more concerning than a model that has lower perplexity, but is incapable of ideating beyond narrow constraints imposed by its training data.

gLMs might also improve our understanding of higher-order genomic architecture, further improving the ability to perform complex pathogen engineering. Thinking further, the broad application of gLMs towards autonomous exploration of the biological landscape through model development feedback loops with limited human checkpoints could have unanticipated risks \citep{wang_safeguards}.

Still, limiting gLM capabilities related to viruses may also limit advances in beneficial public health applications, for example, in antigen design for vaccines, where rapid development is particularly relevant in pandemic and epidemic scenarios \citep{kraemer_artificial_2025}.

\subsection{“Rule-out” evaluations for harm}
We propose the development of evaluations that are explicitly designed to rule out potential harms from gLMs. In this context, ruling out means demonstrating with sufficient confidence that particular harms are absent or acceptably unlikely, whereas ruling in means confirming that such harms are present, which is currently usually conducted through red-teaming. Evaluations with high sensitivity are therefore critical for safety assurance. While more research is needed on how such evaluations would be conducted exactly, to promote usability and scalability, these evaluations would ideally be reproducible across gLMs rather than being custom-built for individual models.

\subsection{Future risk trajectories}

When considering the safety and security of gLMs, researchers should also anticipate future trends in AI development, including multi-modal models \citep{fallahpour_bioreason_2025} and agentic systems \citep{openai_gpt-5_2025}, which introduce new categories of misuse or misapplication risks beyond those studied to date. In the modern AI ecosystem, outputs from one tool can readily become inputs to another. The greatest value will likely be unlocked by highly integrated systems, where model inputs and outputs are seamlessly linked. However, this interconnectedness creates safety and security challenges. Risk assessment of an individual tool is often insufficient to capture risks that emerge across the wider ecosystem (the “daisy-chain” problem). Further innovation in developing such evaluations will be required.

\subsection{Risk mitigation measure applicability to gLMs}
Our evaluations assess the efficacy of the data exclusion risk mitigation approach for gLMs, which falls in the broader risk mitigation category of "capability limitation" \citep{fmf_mitigations_2025}. Capability limitation is a controversial mitigation measure approach, given that many misuse-enabling capabilities are dual-use and can also have desirable applications. For example, the generation of functional viral genomes could also be useful for gene therapy research. 

Aside from capability limitation, there are several other approaches for improving the safety and security of gLMs that could be further explored and developed, such as detection and intervention measures or access controls. LLM developers have developed various technical risk mitigation measures associated with their models, such as refusal training, constitutional classifiers, or false learning \citep{fmf_mitigations_2025}. Researchers are only beginning to explore how and to what degree these risk mitigation measures can be applied to both narrow biological tools as well as biological foundation models. Not all available mitigation measures will be appropriate for gLMs, and novel technical methods will likely be needed \citep{wang_call_2025}. In particular, risk mitigation measures for open-source models should be explored more as open models are prevalent in computational biology \citep{tamirisa_tamper-resistant_2025}.

\subsection{Towards gLM safety frameworks}

Scientists have generally long recognized the importance of frameworks to assess the risks and benefits of research involving human subjects and animals. Within AI development, safety frameworks have also emerged as a critical tool for frontier AI developers, aiding in the assessment and management of a diversity of risks from advanced AI systems \citep{forum_introducing_2025}. Given the rapid advancements in the field, we wish to highlight the importance of developing similar safety frameworks for gLMs, so that their development and deployment can advance responsibly, the manifold benefits can be harnessed, and misuse risks can be effectively mitigated.

\section{Conclusion}

We demonstrate that, in the case of open-source models, it is possible to circumvent data exclusion safeguards via fine-tuning with human-infecting virus sequences. If sensitive pathogen data such as this is publicly accessible, it can be used to fine-tune an openly available gLM, thereby rescuing misuse-enabling capabilities. Importantly, the rescued performance did not match that of a narrow, purpose-built tool for functional immune escape prediction tasks. Therefore, while sensitive data exclusion raises the bar for misuse, it is susceptible to circumvention. Other risk mitigation measures for gLMs will be required in addition. To ensure safe and responsible development and deployment of this class of powerful biological AI models, further work is needed to develop a taxonomy of misuse-enabling capabilities and a corresponding toolkit of implementable gLM-specific risk evaluation and mitigation measures. Ultimately, these efforts should contribute towards comprehensive safety frameworks for gLMs that developers and deployers can implement to manage risks.

\newpage

\begin{ack}
The authors thank Garyk Brixi and Ishan Mukherjee for assistance with fine-tuning Evo 2. The authors also thank two anonymous peer reviewers and Coleman Breen for helpful feedback.
\end{ack}

Funding (direct support): JRMB—Open Philanthropy; MSH—Horizon Institute for Public Service (scholarship); AM—Open Philanthropy (Ph.D.); THB—NIH NCATS (1UM1TR004921), AHRQ (R01HS024096), NIH NLM (R01LM013362); OMC-New College Todd–Bird Junior Research Fellowship; MRC Fellowship MR/Y010078/1; JP—Chan Zuckerberg Initiative; Open Philanthropy.

Competing interests: THB—royalties or licenses (Coursera, AI in Healthcare); consulting fees (PAUL HARTMANN AG; Grai-Matter; Roche); stock or stock options (Verantos Inc.; Grai-Matter); advisory board (AtheloHealth). OMC—consulting fees (Pelago Biosciences; Faculty.ai; MarketCast); scientific advisory board member (Evolvere Biosciences). JP-consulting fees (Chan Zuckerberg Initiative). All others had no competing interests to declare. 

No funder had a role in the research or decision to publish.

\bibliographystyle{plainnat}
\bibliography{references}





\end{document}